\pdfoutput=1
\documentclass[11pt]{article}
\usepackage[final]{acl}

\usepackage{times}
\usepackage{latexsym}
\usepackage[T1]{fontenc}
\usepackage[utf8]{inputenc}
\usepackage{microtype}
\usepackage{longtable}
\usepackage[table,xcdraw]{xcolor}
\usepackage{booktabs}
\usepackage{float}
\usepackage{outlines}
\usepackage{amsmath}
\usepackage{hyperref}
\usepackage{dblfloatfix}
\usepackage{pifont}
\usepackage{amsfonts}
\usepackage{array}
\usepackage{multirow}
\usepackage{todonotes}
\usepackage{inconsolata}
\usepackage{graphicx}
\usepackage{csquotes}  
\usepackage{tcolorbox}
\usepackage{enumitem}

\newtcolorbox{mybox}[1]{
  colback=cyan!5!white,
  colframe=cyan!75!black,
  fonttitle=\bfseries,
  title=#1,
  left=3pt,
  right=3pt,
  top=2pt,
  bottom=2pt
}

\title{Foundations of LLM Knowledge Materialization:\\
Termination, Reproducibility, Robustness}

\author{
  Luca Giordano \quad \quad Simon Razniewski \\
  ScaDS.AI Dresden/Leipzig \& TU Dresden, Germany \\
  \texttt{\{luca.giordano,simon.razniewski\}@tu-dresden.de}
}

\begin{document}
\maketitle
\begin{abstract}
Large Language Models (LLMs) encode substantial factual knowledge, yet measuring and systematizing this knowledge remains challenging. Converting it into structured format—for example through recursive extraction approaches such as the GPTKB methodology \cite{hu2025enabling}—is still underexplored. Key open questions include whether such extraction can terminate, whether its outputs are reproducible, and how robust they are to variations.

We systematically study LLM knowledge materialization using \textit{miniGPTKBs} (domain-specific, tractable subcrawls), analyzing termination, reproducibility, and robustness across three categories of metrics: yield, lexical similarity, and semantic similarity. We experiment with four variations (seed, language, randomness, model) and three illustrative domains (from history, entertainment, and finance).

Our findings show (i) high termination rates, though model-dependent; (ii) mixed reproducibility; and (iii) robustness that varies by perturbation type—high for seeds and temperature, lower for languages and models. These results suggest that LLM knowledge materialization can reliably surface core knowledge, while also revealing important limitations.

\end{abstract}

\section{Introduction}

\medskip
\begin{quote}
\itshape
This was the man to whom all things were known; this was the king who knew the countries of the world.

\hfill --- \textit{The Epic of Gilgamesh}, \textsc{Prologue}
\end{quote}

\begin{figure}[t]
    \centering
    \includegraphics[width=1.05\linewidth]{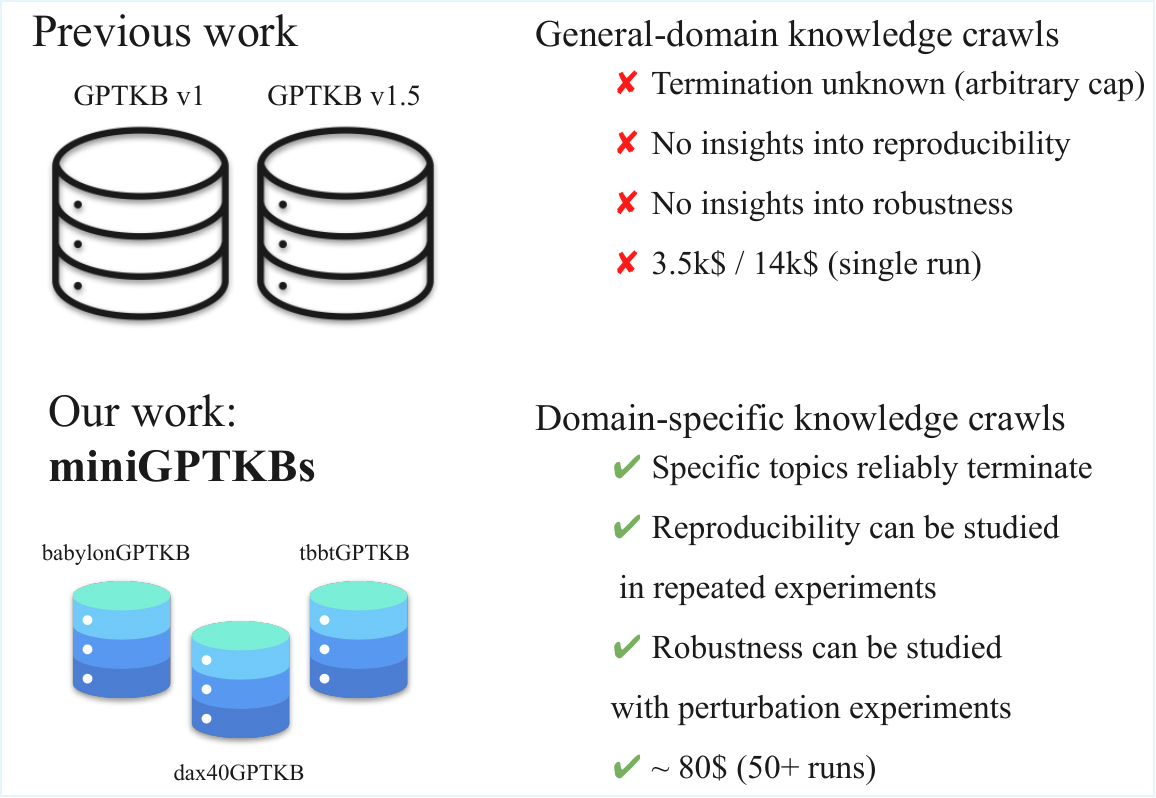}
    \caption{Overview of our approach.}
    \label{fig:teaser}
\end{figure}

\paragraph{Motivation}
Large Language Models (LLMs) have revitalized Artificial Intelligence (AI) and Natural Language Processing (NLP), and impacted many other domains both inside Computer Science \citep{pramanick2024transforming} and beyond, e.g.\ in education, healthcare, and finance \citep{bharathi2024analysis}. During pre-training, LLMs encode in their parameters vast amounts of factual knowledge \citep{bubeck2023sparks}, which can then be implicitly leveraged for a wide range of downstream tasks. However, they also exhibit shortcomings, including lack of coherence, biases, hallucinations, and factual errors \citep{holtzman2020curious, ji2023survey, berglund2024reversal}. Moreover, this factual knowledge is difficult to measure, examine, interpret, and ultimately retrieve. On-demand prompting remains the standard but imperfect means of accessing it. As \citet{cohen2023crawling} note, it is desirable to obtain interpretable, transparent representations of this knowledge and interact with it through tools such as query languages.

Benchmark-based ``knowledge probing'' \cite{petroni2019language} relies on pre-designed cloze or Q\&A tasks, which, as pointed out by \citet{hu2025enabling}, suffers from availability bias \citep{kahnemann}, i.e.\ it cannot reveal knowledge outside the benchmark’s scope.

A more promising approach was proposed by \citet{cohen2023crawling} and popularized by \citet{hu2025enabling, hu2025gptkb}, termed ``knowledge crawling'', i.e.\ prompt-based, recursive, and open-ended knowledge extraction that stores results in a structured knowledge base (KB) of entities and relations, linked in statements in the form of s-p-o-triples. KBs are human-readable, interpretable, interconnected, comparable and easily queried or updated, making them a natural choice for explicit knowledge representation \citep{cohen2023crawling}.

The GPTKB approach at LLM knowledge materialization by \citet{hu2025enabling, hu2025gptkb} represents a prototypical execution at impressive scale, but is still arbitrarily stopped at 100 M triples, and fundamental methodological questions remain unanswered. The present work asks whether the GPTKB approach captures the essence of an LLM's factual knowledge.
\begin{mybox}{Research Questions}
    \begin{itemize}[leftmargin=3em]
    \setlength\itemsep{0pt}
    \item[\textbf{RQ1:}] Can LLM knowledge materialization reach \textbf{termination}, or is it prone to endless hallucination?
    \item[\textbf{RQ2:}] Given that LLMs are nondeterministic, how \textbf{reproducible} is LLM knowledge materialization?
    \item[\textbf{RQ3:}] To what extent are LLM knowledge materializations \textbf{robust} to design choices?
\end{itemize}
\end{mybox}

\noindent
\textbf{Contributions.} Our core contributions are:
\begin{enumerate}
    \item We introduce the concept of \textit{miniGPTKBs}, i.e.\ domain-specific, tractable subcrawls of the whole LLM knowledge, such as the three illustrative examples of {babylonGPTKB} (history), {tbbtGPTKB}\footnote{TBBT = The Big Bang Theory, an American TV sitcom.} (entertainment), and {dax40GPTKB} (finance);
    \item We provide the first proof that the GPTKB approach at LLM knowledge materialization can terminate;
    \item In terms of reproducibility, we observe mixed signals across runs, with low lexical similarity, intermediate semantic similarity, but high similarity in quantitative yield;
    \item In terms of robustness, we find high robustness to different seed entities and temperature values, but lower robustness to different languages and models.
\end{enumerate}
To address the mixed stability (reproducibility \& robustness), we also show that ensembling can significantly enhance the stability of the output. Altogether, our results confirm that, especially with ensembling, the GPTKB methodology provides a stable view of the core of a specific LLM's factual knowledge, with the two main sources of instability being a multilingual setting and the properties/capabilities of specific LLMs.

Data and code are available under the CC-BY 4.0 license on this work's GitHub repository.\footnote{\href{https://github.com/Knowledge-aware-AI/miniGPTKB}{https://github.com/Knowledge-aware-AI/miniGPTKB}}

\section{Related Work}

\paragraph{Knowledge extraction from LLMs}
\citet{petroni2019language} first proposed treating LMs as knowledge bases with the LAMA benchmark, using cloze-style prompts derived from \textit{(s, p, o)} triples, and \citet{kassner-etal-2021-multilingual} extended it to 53 languages. To address the ad-hoc nature of hand-crafted prompts, some works investigated automatic prompt optimization \cite{jiang2020how, shin-etal-2020-autoprompt, wu2024towards}.

\citet{swamy2021interpreting} analyzed knowledge acquisition during training with cloze probing, showing correlations between model knowledge and size. More recently, \citet{cohen2023crawling} extracted KGs from LMs via a “crawling” method starting from seed entities. Similarly, \citet{parovic-etal-2025-generating} generated domain-specific KGs using LLM-produced schemas and Chain-of-Verification prompting.

\paragraph{Reproducibility of LLM outputs} \citet{zheng2024reliable} studied LLM reproducibility, termed 'consistency', evaluating 26 models and finding that more factual models tend to give consistent—sometimes incorrect—responses.
\citet{rajan-etal-2024-knowledge} introduced KonTest, an automated framework to detect inconsistencies in LLM world knowledge using semantically equivalent queries.

\paragraph{Robustness of LLM outputs} \citet{deng2024multilingual} showed that ChatGPT and GPT-4 struggle with language variation, being significantly more likely to produce harmful content when prompted in low-resource languages instead of higher-resource ones.
\citet{lim2025language} investigated the phenomenon of cross-lingual transfer in LLMs i.e.\ their ability to generalize knowledge from one language to another, and suggested that LLMs’ inconsistent outputs across languages may result from relying on language-specific subspaces rather than a shared multilingual semantic space.

\begin{figure*}
    \centering
    \includegraphics[width=1\linewidth]{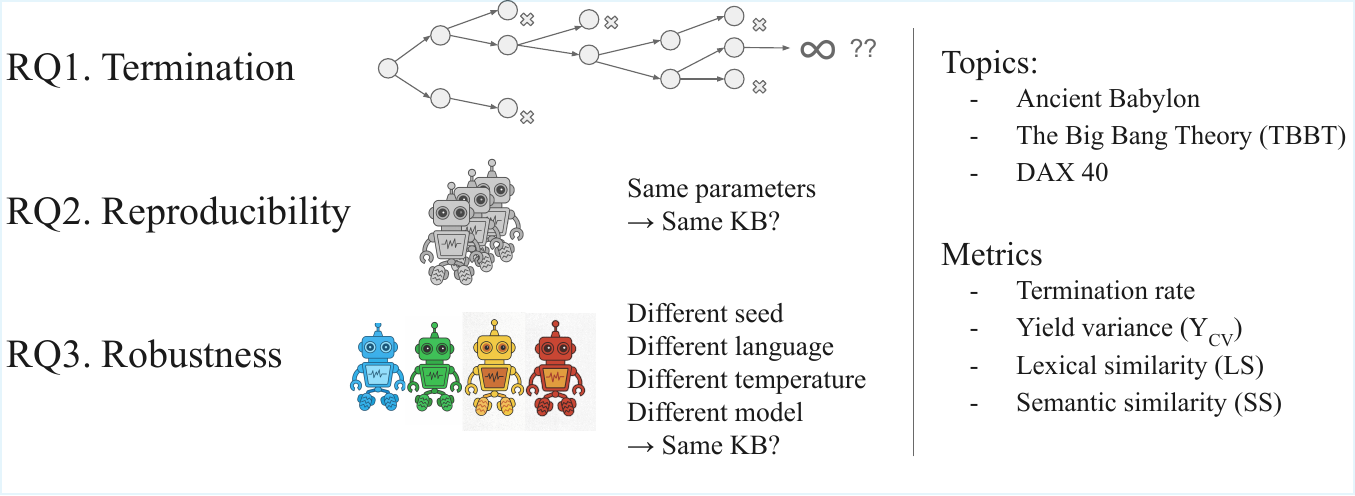}
    \caption{Overview of research questions and methodology.}
    \label{fig:enter-label}
\end{figure*}

\section{GPTKB Method}
The GPTKB approach at LLM knowledge materialization introduced by \citet{hu2025enabling}, further refined in \citet{hu2025gptkb}, and inspired by \citet{cohen2023crawling}, is a form of LLM knowledge crawling, and can be understood as a prompt-based, recursive, and open-ended knowledge extraction process that stores results in a structured format, such as a knowledge base (KB).

At a high level, the workflow consists of two fundamental phases, termed Knowledge Elicitation and Knowledge Consolidation.\footnote{Here we focus only on the first phase. For further details on the second phase, refer to  \cite{hu2025enabling}.}
The process starts from a single arbitrary seed entity, and the LLM is prompted to return knowledge about this subject in the form of \textit{(s, p, o)} triples. Using the model itself for Named Entity Recognition (NER), new named entities are iteratively identified among the objects of these triples. These objects become new subjects and the graph is further expanded recursively.

\citet{hu2025enabling} executed a single run with \textbf{GPT-4o-mini}, only \textbf{in English}, with a \textbf{single arbitrary seed entity} (\textit{Vannevar Bush}) and the \textbf{temperature set to 0}. The elicitation phase was artificially capped at BFS-depth 10, yielding 101M triples for 2.9M entities, 567K relations and 4,715 classes. The process took 27 hours and cost \$3,500 in total. In follow-up work, over \$14k were spent on a similar-sized run with GPT-4.1 \cite{hu2025gptkb}. Although the choice to execute a single, capped run is understandable considering the resources needed, no proof of termination could be provided, and reproducibility and robustness remain open, too. The methodology and experiments in this work aim at shedding light on these questions.

\section{Methodology}
This work adopts the methodology of the GPTKB approach and expands on it to test for termination, reproducibility, and robustness of the knowledge materialization process through \textit{miniGPTKBs}. To compare the resulting KBs of different runs under different experimental settings, we identify their main \textbf{structural categories}: named entities, literals, predicates, classes, and triples.

We test across multiple \textbf{topics} to assess validity across domains. Suitable topics for \textit{miniGPTKBs} in the context of termination and stability testing balance specificity (for tractability), heterogeneity (for varied entity types), and independence from the LLM’s cutoff date. We select three: ancient Babylon (history), The Big Bang Theory (entertainment), and the DAX 40 German stock market index (finance). All experiments use the prompts of \citet{hu2025enabling}, slightly adapted to contain the output of the model within the topics. The full prompts used are found in Appendix~\ref{app:prompts}.

\subsection{Metrics}
Termination is easily reported as a binary observation. To evaluate reproducibility and robustness per structural category across runs, we define three types of metrics:
\begin{enumerate}
    \item \textbf{Yield (Y)}: the count of elements per structural category. Where appropriate, we report its coefficient of variation $Y_{CV} = \frac{\sigma}{\mu}$ to measure variability relative to the mean.
    \item \textbf{Lexical Similarity (LS)}: superficial, string-based similarity, quantified in the following metric computed pair-wise:
    \begin{enumerate}
        \item Average Jaccard similarity (Equation 1 in Appendix~\ref{app:metrics})
    \end{enumerate}
    \item \textbf{Semantic Similarity (SS)}: deeper, embedding-based similarity, quantified in the following two metrics computed pair-wise:
    \begin{enumerate}
        \item Average Cosine-based Hausdorff similarity (Equations 2 and 3 in Appendix~\ref{app:metrics})\footnote{The embedding model used is sentence-transformers/all-MiniLM-L6-v2, while for multilingual comparisons sentence-transformers/paraphrase-multilingual-mpnet-base-v2.}
        \item Bidirectional average percentage of semantic matches (Equations \ref{eq:4} and \ref{eq:5} in Appendix~\ref{app:metrics})
    \end{enumerate}
\end{enumerate}
For example, suppose two runs yield 2 and 3 entities: $r_1=\{\textit{Hammurabi, Marduk Temple, Nebuchadnezzar}\}$ and $r_2=\{\textit{Hammurabi, Temple of Marduk}\}$. The yield variance is Y$_{CV}$ $\tfrac{0.5}{2.5}=0.2$. They share one exact entity, giving LS $=$ 0.25. Using entity embeddings, the average cosine similarity SS1 is 0.83. With threshold $\tau=0.95$, $\frac{2}{3}$ of $r_1$ and both of $r_2$ match, for an average match SS2 of 83.3\%.

Additionally, we want to investigate the influence of entity popularity on stability. Here we define popularity simply as the entity's count of statements on Wikidata \cite{vrandecic2014wikidata}. We partition the named entities of each run in 5 popularity buckets: 4 for the quartiles of the entity list sorted by popularity, and one containing the entities not found on Wikidata. We then compute the metrics pair-wise separately for each bucket within each setting against the full sets of entities of the other runs in turn. 

\subsection{Reproducibility Analysis}
To test for reproducibility, we define a \textbf{base setting}, and execute 30 runs divided by the three topics without altering any variable. Mirroring \citet{hu2025enabling}, we define the base setting's parameters as:
\begin{outline}
    \1 Seed entity:
        \2 babylonGPTKB : \textit{Hammurabi}
        \2 tbbtGPTKB : \textit{Sheldon Cooper}
        \2 dax40GPTKB : \textit{DAX 40}
    \1 Prompt language : English
    \1 Model randomness (temperature) : 0.0
    \1 Model : GPT-4.1-mini
\end{outline}

\subsection{Robustness Analysis}
\label{robustness_methodology}
In this experiment we focus on the topic of Ancient Babylon. We evaluate robustness across \textbf{four dimensions}, 
executing 10 runs with different seeds, 10 runs using different languages, 10 runs with randomness introduced (temperature), and 12 runs across 6 LLMs, two for each:
\begin{itemize}
    \item Seed entity : [\textit{Hammurabi, Tashmetum, Belshazzar, Kudurru, Akitu, Hanging Gardens of Babylon, Shar-Kali-Sharri, Akkadian language, Enuma Elish, Tamkaru}]
    \item Prompt language : [English, Italian, German, Russian, Spanish, French, Turkish, Swedish, Arabic, Polish]
    \item Model randomness (temperature) : 1.0
    \item Model : [GPT-4.1-mini, gpt-oss-120b, Llama 4 Scout, Llama 3.3 70b, DeepSeek-R1, Teuken 7b Instruct]
\end{itemize}
We select a varied set of seed entities, ranging from popular kings, concepts, and locations to niche deities and objects. Regarding the language selection, we select languages from the Germanic, Romance, Slavic, Turkic and Semitic language families. We also consider a variation where we machine-translate other output to English.

For dimension 4, we compare 6 different LLMs: GPT-4.1-mini \cite{openai2025gpt41}, gpt-oss-120b \cite{openai2025gptoss120bgptoss20bmodel}, Llama 4 Scout \cite{meta2025llama4}, Llama 3.3 70b \cite{grattafiori2024llama}, DeepSeek-R1 \cite{guo2025deepseek}, and Teuken 7b Instruct \cite{ali2024teuken}. The hardware specifications used for running the local models can be found in Appendix~\ref{app:hardware_specs}.

\section{Results and Discussion}
We report results for termination, reproducibility, and robustness across all runs and settings. All experiments were conducted between June and July 2025, and the total cost for OpenAI's BatchAPI calls including trial runs amounts to $\sim$80\$.

\subsection{Termination}
\label{termination}

\begin{table*}[t]
\setlength{\tabcolsep}{8pt} 
\centering
\resizebox{2.0\columnwidth}{!}{%
\begin{tabular}{c|c|r|r|r|r|r|r}
\hline
\textbf{Model} & \textbf{Setting} & \textbf{Term.} & \textbf{Avg. runtime} & \textbf{Avg. NEs} & \textbf{Avg. pred.} & \textbf{Avg. triples} & \textbf{Avg. layers} \\
\hline
\multirow{6}{*}{GPT-4.1-mini} 
 & Base (babylon) & 10/10 & 3h & 1669.5 & 1163.4 & 11,670.8 & 20\\
 & Base (tbbt) & 10/10 & 2.5h & 163.1 & 233.1 & 1137.2 & 11\\
 & Base (dax40) & 10/10 & 2h & 273.7 & 240.7 & 1655.5 & 8\\
 & Seed & 10/10 & 3h & 1683.4 & 1166.8 & 11,870.7 & 20\\
 & Language & 7/10 & 11h & 4205.3 & 3911.9 & 31,975.6 & 19 \\
 & Temperature & 10/10 & 3h & 1705.5 & 1149.5 & 11,919.7 & 20\\
\hline
gpt-oss-120b     & \multirow{5}{*}{Base (babylon)} & 2/2 & 1h & 960 & 521.5 & 5689 & 23\\
Llama 3.3 70b    &                                & 2/2 & 1h & 737 & 380.5 & 4131 & 21\\
Llama 4 Scout    &                                & 0/2 & 96h & 440K & 817 & 2.5M & 25\\
DeepSeek-R1      &                                & 0/2 & 96h & 190K & 8583 & 1.2M & 24\\
Teuken 7b Instr. &                                & 0/2 & 96h & 5199 & 560 & 16,550 & 6\\
\hline
\end{tabular}}
\caption{Termination rates across different models and settings, along with average runtime, average yield of NEs and triples, and average deepest BFS-layer.}
\label{tab:termination_models}
\end{table*}

Results for termination can be found in Table \ref{tab:termination_models}. Overall termination was high across all three topics. All 10 runs of each base setting and of the seed/temperature perturbations reliably terminated with similar runtime ($\sim$3 hours), yield ($\sim$1700 NEs and $\sim$1160 predicates), and BFS-depth ($\sim$20). However, 3 out of the 10 runs of the language perturbation setting (i.e.\ Italian, German, French) did not terminate organically and had to be arbitrarily suspended to nonetheless allow for comparison. These runs continued for 96h without completing (vs.\ $\sim$3h for other babylonGPTKB settings, and less for other topics). The main pattern we notice related to the lack of termination for these settings is the endless generation of ever so slightly different synonyms of existing entities (e.g.\ Italian \textit{governo di Elam} -> \textit{governo di Elam durante il periodo babilonese}\footnote{\textit{Elam's government} -> \textit{Elam's government during the babylonian period}}), but we cannot exclude a potential future termination at some point. It may seem that the lack of termination relates to the over-representation of these high-resource languages in web and LLM training data. However, (1) other high-resource languages (e.g.\, English, Russian, Spanish) still terminated reliably, and (2) the correlation between named entities per language and web distribution (OSCAR token counts \cite{OrtizSuarezSagotRomary2019}) is weak and not statistically significant (Figure \ref{fig:oscar} in Appendix~\ref{app:additional_visualizations}).
\paragraph{Local models} As shown in Table \ref{tab:termination_models}, termination could be observed for two of the five local models i.e.\ gpt-oss-120b and Llama 3.3 70b, which terminated within a tractable runtime of around 1h, a reasonable average yield (e.g.\, 960 NEs and 5689 triples for gpt-oss-120b and 737 NEs and 4131 triples for Llama 3.3) and BFS-depth, comparable with GPT-4.1-mini. On the other hand, over 96h of runtime, despite the similar BFS-depth, we observe no termination for the remaining three models, which either generated (and kept generating) disproportionally more (mostly hallucinated) triples than GPT-4.1-mini (e.g.\ 2.5M for Llama 4 Scout and 1.2M for DeepSeek-R1 vs.\ 11K for GPT), or, in the case of Teuken, generated incorrect output formats which would repeatedly break the extraction, making its termination problematic. Analyzing the triples produced by these three models, we identify four main types of undesired behavior: 1) (often invented) Wikidata-style Q identifiers as entities (e.g.\ \textit{Q768509} is an entity generated by Llama 4), 2) questionable NER classifications (e.g.\ \textit{section of Babylon} is classified as a NE by DeepSeek-R1), 3) off-topic entities and triples (e.g.\ \textit{Iraq's Central Bank} is an entity generated by Llama 4, but it was established in 1947, long after the Babylonian Empire), and 4) repetitive loops in entity names (e.g.\, \textit{Nabu-mukin-zeri-mu-mu-mu}… generated by Teuken), where the model endlessly repeats a syllable which at each iteration counts as a new entity. We suspect that the lack of termination for these models and the undesired behaviors mentioned above might be related to the phenomena of \textit{text degeneration} i.e.\ ``text that is bland, incoherent, or stuck in repetitive loops'' \cite{holtzman2020curious} and \textit{verbosity compensation} i.e.\ the tendency observed in LLMs to compensate uncertainty and lack of knowledge with a verbose output \cite{zhang2024verbosity}. In particular, \citet{zhang2024verbosity} observe the latter phenomenon to be prevalent in open source LLMs, which might partially explain the undesired behaviors observed in the local models tested. This leads us to suspect that capability for termination might be model-dependent, and perhaps to some degree related to the model's hallucination rate. 

\begin{mybox}{Result 1: Termination}
    LLM knowledge materialization reliably terminates across 30 runs and 3 topics with GPT-4.1-mini, gpt-oss-120b and Llama 3.3. No termination with three out of ten languages and Llama 4, DeepSeek-R1 and Teuken. 
\end{mybox}

\begin{table}[t]
\centering
\renewcommand{\arraystretch}{1.05}
\setlength{\tabcolsep}{5pt}

\begin{tabular}{l c c c c}
\toprule
 & \textbf{$\downarrow$Y$_{CV}$} & $\uparrow$\textbf{LS} & $\uparrow$\textbf{SS1} & $\uparrow$\textbf{SS2} \\
\midrule
\textit{Baseline}     & - & 0.0 & 0.44 & 0.0\% \\
\hline
babylonGPTKB & 0.12 & 0.33 & 0.89 & 58.3\% \\
tbbtGPTKB    & 0.18  & 0.41 & 0.86 & 61.3\% \\
dax40GPTKB   & 0.13  & 0.51 & 0.89 & 69.0\% \\
\bottomrule
\end{tabular}
\caption{Average of metrics for NEs across \textbf{reproducibility} topics. Y = Yield (reported here as coefficient of variation $CV=\frac{\sigma}{\mu}$), LS = Jaccard similarity, SS1 = Cosine-based Hausdorff similarity, SS2 = bidirectional average percentage of semantic matches. The baseline is a comparison of 3 runs for intentionally unrelated domains, to indicate the calibration of the metrics.}
\label{tab:NE_sim_reprod}
\end{table}

\subsection{Reproducibility}
\label{reproducibility}
We observe mixed signals of reproducibility across the thirty runs divided by the three topics of the base setting.

\paragraph{Yield} Yield shows low coefficients of variation, reported in Table \ref{tab:NE_sim_reprod} for NEs and Table \ref{tab:other_sim_reprod} for other structural categories (Appendix \ref{app:sim}). The lower the coefficient, the higher the reproducibility, and this applies to all three topics. Furthermore, as shown in Table \ref{tab:combined_yield}, the yield metric for all topics and structural categories is very similar among runs also in terms of absolute values, shown in means and low standard deviations. Thus, rerunning our base setting ten times generates KBs with approximately the same yield per structural category. The narrowest topic is The Big Bang Theory (163.1 ± 29.4 average NEs), followed by DAX 40 (273.7 ± 37.7 average NEs) and Ancient Babylon (1669.5 ± 202.1 average NEs), which is much broader.
\paragraph{Lexical Similarity} As reported in Table \ref{tab:NE_sim_reprod} for NEs and Table \ref{tab:other_sim_reprod} for other categories, in babylonGPTKB we observe a relatively low exact-match lexical similarity, which suggests that only around \(\frac{1}{3}\) of elements matches exactly in other runs. For example, the lexical similarity for named entities is 0.33±0.02, for predicates it is the highest at 0.43±0.03, for classes 0.39±0.03, while for literals the lowest at 0.26±0.02. Lexical similarity in tbbtGPTKB and dax40GPTKB is higher but still modest, for named entities at 0.41±0.04 and 0.51±0.04 respectively.
\paragraph{Semantic Similarity} Also the semantic similarity metrics are reported in Table \ref{tab:NE_sim_reprod} and Table \ref{tab:other_sim_reprod}. Going beyond the stricter exact-matching, the cosine similarity e.g.\ for named entities and literals in babylonGPTKB is 0.89, for predicates 0.88 and for classes even 0.91. Most interestingly, in babylonGPTKB the bidirectional average percentage of semantic matches (similarity >=0.95) ranges across structural categories from around \(\frac{1}{2}\) to \(\frac{2}{3}\) of all pairs (from 53.09\% for literals to 66.9\% for classes), while even reaching 75.7\% for classes in tbbtGPTKB. Thus, reruns of babylonGPTKB's base setting often yields semantic matches, and for the two other topics even more.
\paragraph{Correlation with entity popularity} We observe in all topics a consistent positive correlation between reproducibility and entity popularity (Figure \ref{fig:popularity} in Appendix~\ref{app:popularity}). In the base setting, both the lexical and semantic similarity metrics for named entities increase linearly with entity popularity. The lowest lexical similarity across runs (e.g.\ in babylonGPTKB 0.22) can be observed for the bucket of entities not found on Wikidata (likely representing the long-tail knowledge, which drags down the average), and it gradually increases, reaching 0.51 for the highest popularity quartile in babylonGPTKB. The same trend is observed with semantic matches, ranging from 48.7\% to 72.3\%.\footnote{Noteworthy here is the limitation of the cosine similarity, which tends to flatten differences in the shared embedding space. Therefore, here we focus on the semantic match metric, which manages to capture these differences.} In the other topics, there seems to be a sharp increase in the first buckets (e.g.\ in dax40GPTKB from 50.07\% of entities not found on Wikidata to 81.1\% in the third quartile) followed by a slight dip in the last quartile (78\%).

Overall, over repeated runs with the same model, with the GPTKB approach at LLM knowledge materialization one can expect the model to generate KBs with almost the same yield per structural category, containing at least around \(\frac{1}{3}\) of the exact same elements (higher when considering other topics or only popular entities) and at least half of them (or at least \(\frac{2}{3}\) for popular entities) having at least a very close semantic match in the other runs.

\begin{mybox}{Result 2: Reproducibility}
    Repeated runs of LLM knowledge materialization show high reproducibility in terms of output size (yield), moderate lexical similarity, and fairly high semantic similarity. The knowledge extracted is consistent but variably expressed.
\end{mybox}

\begin{table}[t]
\centering
\renewcommand{\arraystretch}{1.05}
\setlength{\tabcolsep}{5.5pt}

\begin{tabular}{l c c c c}
\toprule
 & \textbf{$\downarrow$Y$_{CV}$} & \textbf{$\uparrow$LS} & \textbf{$\uparrow$SS1} & \textbf{$\uparrow$SS2} \\
\midrule
\textit{Base setting}       & 0.12 & 0.33 & 0.89 & 58.3\% \\
\hline
Seed               & 0.10 & 0.33 & 0.89 & 57.9\% \\
Language           & 0.90 & 0.01 & 0.85 & 21.8\% \\
Language (tr.) & 0.90 & 0.06 & 0.82 & 28.1\% \\
Temperature        & 0.08 & 0.32 & 0.89 & 57.2\% \\
Model              & 0.35 & 0.27 & 0.85 & 49.7\% \\
\bottomrule
\end{tabular}
\caption{Average of metrics for NEs across \textbf{robustness} settings. The values reported for the model setting refer only to the models that terminated. Y = Yield (reported here as coefficient of variation $CV = \frac{\sigma}{\mu}$), LS = Jaccard similarity, SS1 = Cosine-based Hausdorff similarity, SS2 = bidirectional average percentage of semantic matches.}
\label{tab:NE_sim_robust}
\end{table}

\begin{table*}[ht]
\centering
\renewcommand{\arraystretch}{1.2}
\small
\begin{tabular}{lccccc}
  \toprule
  & \textbf{NEs} & \textbf{Literals} & \textbf{Predicates} & \textbf{Classes} & \textbf{Triples} \\
    \midrule
  \multicolumn{6}{l}{\textbf{Reproducibility}} \\
  \midrule
  babylonGPTKB   & 1669.5 ± 202.1 & 3198.8 ± 366.9 & 1163.4 ± 130.1 & 404.3 ± 48.6 & 11,670.8 ± 1454.5 \\
  tbbtGPTKB      & 163.1 ± 29.42  & 298.9 ± 50.01  & 233.1 ± 26.62  & 31.6 ± 4.18  & 1137.2 ± 244.54 \\
  dax40GPTKB     & 273.7 ± 36.73  & 438.4 ± 48.14  & 240.7 ± 25.65  & 30.6 ± 4.10  & 1655.5 ± 193.62 \\
  \midrule
  \multicolumn{6}{l}{\textbf{Robustness}} \\
  \midrule
  Seed           & 1683.4 ± 172.4 & 3221.2 ± 261.1 & 1166.8 ± 85.9  & 416.2 ± 35.3 & 11,870 ± 1201.3 \\
  Language       & 4205.3 ± 3358.7 & 7577.7 ± 6352.5 & 3911.9 ± 3094.2 & 557.8 ± 338.3 & 31,975.6 ± 30,323.5 \\
  Temperature    & 1705.5 ± 146.1 & 3235 ± 291.2 & 1149.5 ± 82.8 & 413.8 ± 39.4 & 11,919.7 ± 1189.9 \\
  Model          & 1122.1 ± 397.5 & 1648.9 ± 1106.9 & 688.4 ± 340.7 & 264.7 ± 101.4 & 7163.6 ± 3249.9\\
  \bottomrule
\end{tabular}
\caption{Yield per structural category (average ± std.) across reproducibility topics and robustness dimensions. The values reported for the model setting refer only to the models that terminated.}
\label{tab:combined_yield}
\end{table*}

\subsection{Robustness}
We observe mixed signals also for robustness, with big differences between settings.

\paragraph{Seed variation}
In terms of yield, as shown in Table \ref{tab:NE_sim_robust} for NEs and in Table \ref{tab:other_sim_robust} for the other structural categories, the seed variation setting is very similar to the base setting, with low coefficients of variation and only marginal differences in yield (reported in Table \ref{tab:combined_yield}). Average yield almost perfectly aligns with the base setting, indicating that changing the seed entity does not significantly impact the yield per structural category in the resulting KB. The same holds for the lexical and semantic similarity metrics: seed variation yields almost identical results to the base setting (e.g. for NEs \, Jaccard similarity at 0.33, cosine-based Hausdorff similarity at 0.89, and semantic matches at 57.9\%). This strongly suggests that the LLM knowledge materialization process is highly robust against seed variation.\footnote{In addition to the robustness analysis described, i.e. 10 runs each with a different seed entity, we also executed 10 unchanged runs for 3 of the seed entities i.e. \textit{Tashmetum}, \textit{Hanging Gardens of Babylon}, and \textit{Tamkaru}, as a middle ground experiment to test reproducibility and robustness at the same time. Results show negligible differences compared to the base setting (approximately 1–3\%) both in terms of yield and the three similarity metrics.}

\paragraph{Language variation}
A very different picture emerges for the language variation setting, which produces inconsistent results across the 10 language runs. Average yield per structural category is more than twice that of other settings (e.g.\ 4205.3 ± 3358.7 for NEs), but with extremely high variation (from 412 for Turkish to 11,759 for Italian). Setting aside the Italian, German, and French runs, which were manually terminated as per Section \ref{termination}, Spanish (2615) and Russian (6016) yielded above the English base setting’s average of 1669.5, while others like Polish (1093), Swedish (1546), Arabic (859), and Turkish (412) yielded below it. This high volatility is reflected in the coefficient of variation (e.g. 0.90 for NEs, 0.79 for predicates), by far the highest among all settings.

For a fair interpretation of the similarity metrics for the language variation setting and comparison to other settings, it is necessary to keep in mind that 1) as stated in Section \ref{robustness_methodology}, we compute the lexical similarity metric on automatic translations into English, which by itself introduces noise, and 2) we use a different (multilingual) embedding model to compute the semantic similarity metrics compared to all other English monolingual settings. Keeping this in consideration, we observe that the lexical similarity is close to 0 across all structural categories (e.g.\, 0.06 for NEs compared to 0.33 in the base setting), likely a consequence of the noisy and indeterministic automatic translation. However, semantic similarity metrics show somewhat higher robustness: focusing on NEs, cosine similarity averages at 0.85 for originals and 0.82 for translations (compared to 0.89 for the base setting). Yet, the proportion of semantic matches is much lower (21.8\% for originals, 28.1\% for translations, compared to 58.3\% for base). Overall, this indicates a low robustness against language variation.

\paragraph{Temperature variation}
The randomness (temperature) variation setting behaves very similarly to the seed variation and base setting. Yield metrics remain stable, with slightly higher counts but negligible variation (coefficient of variation 0.08 vs.\ 0.12 for base). Lexical and semantic similarity metrics also closely align with the base setting: Jaccard similarity of 0.32, cosine similarity of 0.89, and semantic matches of 57.2\%. This confirms that adjusting the temperature parameter has no significant effect on either yield or content similarity of the KB, reinforcing robustness in this dimension.

\paragraph{Model variation}
The model perturbation setting displays intermediate robustness. Its yield coefficient of variation (0.35 for NEs) is much higher than the base, seed, and temperature settings, but still about half that of the language variation. In terms of similarity, both lexical similarity (0.27) and semantic similarity (0.85 cosine similarity, 49.7\% matches) almost align with the better-performing settings, though not as consistently.

\paragraph{Correlation with entity popularity}
For all robustness settings we observe the same positive correlation with entity popularity described in Section \ref{reproducibility} (Figure \ref{fig:popularity} in Appendix~\ref{app:popularity}).

\begin{mybox}{Result 3: Robustness}
LLM knowledge materialization is highly robust against seed and temperature changes, but less robust against language and model variation.
\end{mybox}

\subsection{Ensembling Towards Final KBs}
Inspired by \citet{wang2023self}, we show that simple intersection ensembling across repeated runs improves output stability. For example, with \textit{k}=3, comparing intersections of the base setting of runs 1–3 (689 entities), 4–6 (564 entities), and 7–9 (671 entities) yields much higher similarity than the base setting: average cosine-based Hausdorff similarity rises from 0.89 to \textbf{0.94}, and semantic matches from 58.3\% to \textbf{75\%}. Despite higher computational costs than a single run, we recommend it as a methodologically simple means to a more stable output.

We generate final, reusable versions of the three KBs babylonGPTKB, tbbtGPTKB, and dax40GPTKB by retaining the triples that appear in at least \textit{k} runs of the 10 runs of the base setting for each topic. We experiment with all possible values of $k \in [1,10]$, produce plots for each topic and choose an appropriate value using the elbow method \cite{Thorndike_1953}, which in all three topics gives \textit{k}=3, also resulting in a number of triples closely aligned with the averages observed in Table \ref{tab:combined_yield}. We report in Appendix~\ref{app:ensembling} the plots (Figures \ref{fig:elbow_babylon}, \ref{fig:elbow_tbbt} and \ref{fig:elbow_dax40}) and number of shared triples @k for each topic (Table \ref{tab:elbow}). We release the KBs under the CC-BY 4.0 license in this work's GitHub repository as CSV files, as relational databases, as TTL Turtle serializations, and as an interactive, interlinked collection of HTML files.\footnote{\href{https://github.com/Knowledge-aware-AI/miniGPTKB}{https://github.com/Knowledge-aware-AI/miniGPTKB}}

Finally, we conduct a small-scale accuracy evaluation using human judges by sampling 200 random triples from babylonGPTKB (one half drawn from the first reproducibility run and the second half drawn from the final ensembled KB). Our evaluation shows an accuracy of 93\% and 95\% for the two subsets respectively. Details are reported in Appendix \ref{acc_eval}.

\subsection{Qualitative Insights}
\label{qualitative_insights_main}

In babylonGPTKB, triples generated for three sample entities by GPT-4.1-mini often differ across runs, with some consistently shared and others appearing only in specific runs. Despite these exact-match differences, yield and semantic matching remain stable, indicating that the model’s knowledge extraction is generally robust. 

Additionally, some entities appear in one run but not another—typically due to missing parent entities, absent predicates, or minor variations (e.g.\, spelling). Full qualitative insights and detailed examples are provided in Appendix~\ref{app:qualitative_insights_app}.

\subsection{Result Generalization}

Expensive open-domain runs over \$3k and \$14k have not yet terminated \cite{hu2025enabling,hu2025gptkb}, and no estimate exists for their completion. While our experiments show frequent termination of topic-specific runs, this does not prove open-domain termination, as just a single hallucination-prone subtopic could make it run indefinitely. However, demonstrating that specific topics reliably terminate after limited iterations supports justifying runtime cutoffs for open-domain runs: from Table~\ref{tab:termination_models}, we observe that once a subtopic is reached, it should terminate within ca.\ 20 layers, hence, for a cutoff at layer $n$, all subtopics reached until $n-20$ can be expected to be completed. Moreover, our results on reproducibility and robustness do generalize.

\section{Conclusion}

In this paper we systematically investigated fundamental properties of LLM knowledge materialization: termination, reproducibility, and robustness. With some nuances, our experiments with three domain-specific \textit{miniGPTKBs} prove that the GPTKB method can be used to surface the essence of an LLM's core knowledge.

\section{Limitations}

The first limitation is the choice and breadth of LLMs tested. In reproducibility and robustness, we focused on experiments conducted with GPT-4.1-mini, a closed-source, commercial model, which limits replicability. Nonetheless, this is a conscious choice, for the following reasons: i) matches previous work \cite{hu2025enabling, hu2025gptkb}, ii) ability to reliably process requests in batches via API, iii) structured output feature, and iv) good trade-off between cost and performance. We also tested 5 locally-run open-weights models to investigate the validity of results across different LLM families and architectures, while improving the reproducibility of the experiments. We consider future work to conduct similar experiments across a much larger pool of LLMs, both commercial and open-weights.

The second limitation is that there are other potential dimensions of robustness to investigate, e.g.\ variation in prompt wording or testing a finer gradient for the temperature parameter. While we believe we investigated the most critical and salient robustness dimensions, we consider other potential ones as room for future work.

Finally, in this work, other than the small-scale accuracy evaluation reported in Appendix \ref{acc_eval}, we did not investigate thoroughly the accuracy of the resulting KBs, i.e.\ how many of the facts stated in babylonGPTKB, tbbtGPTKB, and dax40GPTKB are factually true. While we acknowledge the relevance of this question (especially for potential real-world applications of domain-specific \textit{miniGPTKBs}), the evaluation of factual correctness was intentionally left out of the present study, as it was not the primary objective of our experiments and it was already investigated in \citet{hu2025enabling}. Our focus here was to investigate the methodological open questions of termination, reproducibility and robustness of knowledge extraction from LLMs, rather than validating the truthfulness of the resulting KBs.

\subsection*{Acknowledgments}

We thank Shrestha Ghosh, Tuan-Phong Nguyen, and Yujia Hu for valuable discussions and constructive feedback.

This work has been partially supported by the Deutsche Forschungsgemeinschaft (DFG grant number 389792660 as part of the Transregional Collaborative Research Centre TRR 248: Center for Perspicuous Computing, CPEC).

\bibliography{refs.bib}

\clearpage

\appendix
\section*{Appendix}

\section{Prompts}
\label{app:prompts}

In the following, we list the prompts used in our implementation of the GPTKB methodology. The parts in black are taken from \cite{hu2025enabling}, the parts in red represent our adaptation for topic-specific KB construction.

\begin{mybox}{Prompt - Knowledge Elicitation}
    "I want to construct a knowledge graph \textcolor{red}{on the topic of the \{ancient city of Babylon, TV series The Big Bang Theory, DAX 40 Index\}}. Given a subject entity, return all facts that you know for the subject as a list of (subject, predicate, object) triples. The number of facts may be very high, between 50 to 100 or more, for very popular subjects. For less popular subjects, the number of facts can be very low, like 5 or 10. Important:\\
    \begin{itemize}
        \item If you don't know the subject, return an empty list. 
        \item If the subject is not a named entity, return an empty list.
        \item \textcolor{red}{If the subject does not belong to the topic of the \{ancient city of Babylon, TV series The Big Bang Theory, DAX 40 Index\}}, return an empty list.
        \item If the subject is a named entity, include at least one triple where predicate is "instanceOf".
        \item Do not get too wordy. 
        \item Separate several objects into multiple triples with one object."
    \end{itemize}
\end{mybox}
\begin{mybox}{Prompt - NER}
    "I want you to perform named entity recognition (NER) \textcolor{red}{on the topic of the \{ancient city of Babylon, TV series The Big Bang Theory, DAX 40 Index\}}. Your task is to classify if given phrases are \textcolor{red}{topic-relevant} named entities, or not (e.g.\, literals, dates, URLs, verbose phrases...). Each phrase is given to you in a line."
\end{mybox}

\section{Evaluation Metrics}
\label{app:metrics}

\begin{equation}
\text{\textit{AvJSim}} = \frac{2}{n(n-1)} \sum_{1 \leq i < j \leq n} \frac{|S_i \cap S_j|}{|S_i \cup S_j|}
\label{eq:1}
\end{equation}

\begin{multline}
\bar{d}_{A \to B} = \frac{1}{|A|} \sum_{i=1}^{|A|} \min_{j=1}^{|B|} D_{A \to B}[i,j] \\
\bar{d}_{B \to A} = \frac{1}{|B|} \sum_{j=1}^{|B|} \min_{i=1}^{|A|} D_{B \to A}[j,i]
\label{eq:2}
\end{multline}

\begin{multline}
s_{\text{ACH}}(A,B) = 1 - \bar{d}_{avg} = 1 - \frac{\bar{d}_{A \to B} + \bar{d}_{B \to A}}{2}
\label{eq:3}
\end{multline}

\begin{multline}
M_{A \to B} = \sum_{i=1}^{|A|} \mathbb{1}\left[\max_{j=1}^{|B|} S_{A \to B}[i,j] \geq \tau\right] \\
M_{B \to A} = \sum_{j=1}^{|B|} \mathbb{1}\left[\max_{i=1}^{|A|} S_{B \to A}[j,i] \geq \tau\right]
\label{eq:4}
\end{multline}

\begin{multline}
P_{A \to B} = \frac{M_{A \to B}}{|A|} \times 100\% \\
P_{B \to A} = \frac{M_{B \to A}}{|B|} \times 100\% \\
P_{avg} = \frac{P_{A \to B} + P_{B \to A}}{2}
\label{eq:5}
\end{multline}\\
In Equation \ref{eq:1}, $n$ is the total number of sets, $S_i$ and $S_j$ are the $i$-th and $j$-th sets respectively, $|S_i \cap S_j|$ is the cardinality of the intersection of sets $S_i$ and $S_j$, $|S_i \cup S_j|$ is the cardinality of the union of sets $S_i$ and $S_j$, and the summation is over all unique pairs $(i,j)$ with $i < j$.

In Equations \ref{eq:2} and \ref{eq:3}, $\bar{d}_{A \to B}$ is the average cosine-based minimum distance from set $A$ to set $B$, $\bar{d}_{B \to A}$ is the average cosine-based minimum distance from set $B$ to set $A$, $|A|$ and $|B|$ are the cardinalities of sets $A$ and $B$ respectively, $D_{A \to B}[i,j]$ is the cosine distance between the $i$-th element of set $A$ and the $j$-th element of set $B$, and $D_{B \to A}[j,i]$ is the cosine distance between the $j$-th element of set $B$ and the $i$-th element of set $A$.

In Equation \ref{eq:4}, $M_{A \to B}$ is the number of elements in set $A$ that have a match in set $B$, $M_{B \to A}$ is the number of elements in set $B$ that have a match in set $A$, $\mathbb{1}[\cdot]$ is the indicator function, $S_{A \to B}[i,j]$ is the cosine similarity between the $i$-th element of set $A$ and the $j$-th element of set $B$, $S_{B \to A}[j,i]$ is the cosine similarity between the $j$-th element of set $B$ and the $i$-th element of set $A$, and $\tau$ is the similarity threshold. Two elements match when their cosine similarity exceeds the threshold. We set the threshold at 0.95.

Finally, in Equation \ref{eq:5}, $P_{A \to B}$ is the percentage of elements in set $A$ that have a match in set $B$, $P_{B \to A}$ is the percentage of elements in set $B$ that have a match in set $A$, and $P_{avg}$ is bidirectional average percentage of semantic matches.

\section{Correlation with entity popularity}
\label{app:popularity}
\begin{figure*}
    \centering
    \includegraphics[width=0.8\linewidth]{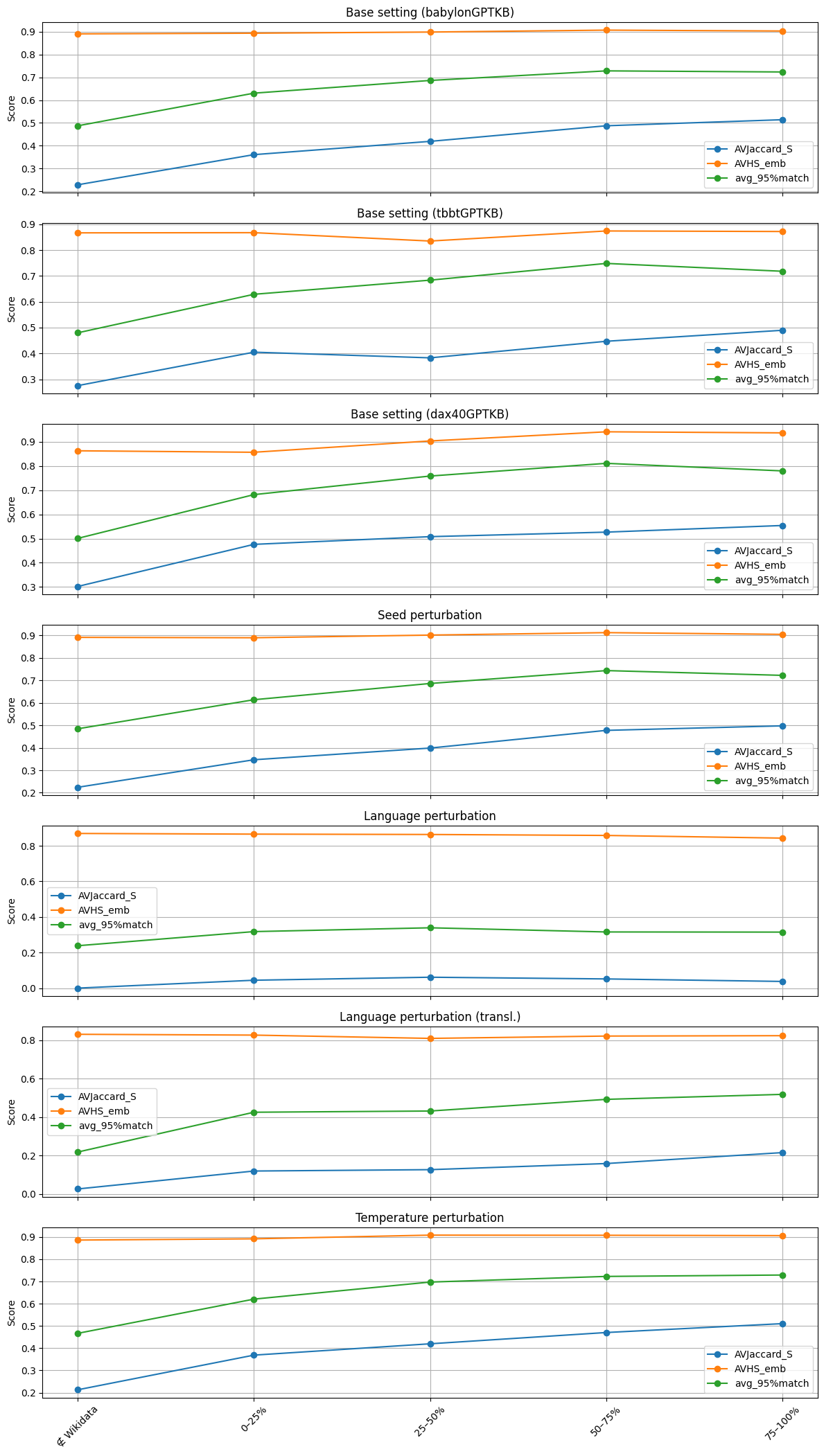}
    \caption{Similarity metrics across entity popularity buckets under the different settings and topics.
}
    \label{fig:popularity}
\end{figure*}

In Figure \ref{fig:popularity} we plot the average for the similarity metrics across all topics and settings (except the model variation) against the five popularity buckets. We observe a general trend of similarity increasing linearly with entity popularity, especially notable in the average Jaccard similarity and in the percentage of semantic matches across runs.

\clearpage

\section{Similarity metrics across other structural categories}
\label{app:sim}

\begin{table}[ht]
\small
\setlength{\tabcolsep}{3pt} 
\centering
\begin{tabular}{l l l c c c c}
\toprule
& \textbf{Category} &  & \textbf{$\downarrow$Y$_{CV}$} & $\uparrow$\textbf{LS} & $\uparrow$\textbf{SS1} & $\uparrow$\textbf{SS2} \\
\midrule

\multirow{3}{*}{babylonGPTKB}
 & Predicates &  & 0.11 & 0.43 & 0.88 & 61.9\% \\
 & Classes    &  & 0.12 & 0.39 & 0.91 & 66.9\% \\
 & Literals   &  & 0.11 & 0.26 & 0.89 & 53\% \\
\midrule

\multirow{3}{*}{tbbtGPTKB}
 & Predicates &  & 0.11 & 0.51 & 0.89 & 69.1\% \\
 & Classes    &  & 0.13 & 0.54 & 0.89 & 75.7\% \\
 & Literals   &  & 0.16 & 0.33 & 0.84 & 55.6\% \\
\midrule

\multirow{3}{*}{dax40GPTKB}
 & Predicates &  & 0.11 & 0.42 & 0.87 & 61.1\% \\
 & Classes    &  & 0.13 & 0.50 & 0.86 & 69.6\% \\
 & Literals   &  & 0.10 & 0.33 & 0.87 & 58.1\% \\
\bottomrule
\end{tabular}

\caption{Average of metrics for predicates, classes, and literals across \textbf{reproducibility} topics. Y = Yield (reported as coefficient of variation $CV=\frac{\sigma}{\mu}$), LS = Jaccard similarity, SS1 = Cosine-based Hausdorff similarity, SS2 = bidirectional average percentage of semantic matches.}
\label{tab:other_sim_reprod}
\end{table}

\begin{table}[ht]
\small
\setlength{\tabcolsep}{3.5pt} 
\centering
\begin{tabular}{l l l c c c c}
\toprule
& \textbf{Category} &  & \textbf{$\downarrow$Y$_{CV}$} & $\uparrow$\textbf{LS} & $\uparrow$\textbf{SS1} & $\uparrow$\textbf{SS2} \\
\midrule

\multirow{3}{*}{Seed}
 & Predicates &  & 0.07 & 0.43 & 0.88 & 61.8\% \\
 & Classes    &  & 0.08 & 0.38 & 0.91 & 66\% \\
 & Literals   &  & 0.08 & 0.25 & 0.89 & 51.5\% \\
\midrule

\multirow{3}{*}{Language}
 & Predicates &  & 0.79 & 0 & 0.84 & 15\% \\
 & Classes    &  & 0.60 & 0 & 0.85 & 26.6\% \\
 & Literals   &  & 0.83 & 0 & 0.86 & 23.2\% \\
\midrule

\multirow{3}{*}{Language (tr.)}
 & Predicates &  & 0.79 & 0.05 & 0.75 & 20.2\% \\
 & Classes    &  & 0.60 & 0.06 & 0.79 & 25.8\% \\
 & Literals   &  & 0.83 & 0.03 & 0.80 & 17.4\% \\
 \midrule

 \multirow{3}{*}{Temperature}
 & Predicates &  & 0.07 & 0.43 & 0.88 & 61.2\% \\
 & Classes    &  & 0.09 & 0.38 & 0.91 & 65.8\% \\
 & Literals   &  & 0.09 & 0.24 & 0.89 & 51.2\% \\
\midrule

\multirow{3}{*}{Model}
 & Predicates &  & 0.49 & 0.10 & 0.69 & 21.5\% \\
 & Classes    &  & 0.38 & 0.06 & 0.72 & 20.2\% \\
 & Literals   &  & 0.67 & 0.02 & 0.62 & 8.9\% \\

\bottomrule
\end{tabular}

\caption{Average of metrics for predicates, classes, and literals across \textbf{robustness} settings. Y = Yield (reported as coefficient of variation $CV=\frac{\sigma}{\mu}$), LS = Jaccard similarity, SS1 = Cosine-based Hausdorff similarity, SS2 = bidirectional average percentage of semantic matches.}
\label{tab:other_sim_robust}
\end{table}

\section{Qualitative Insights}
\label{app:qualitative_insights_app}
In this section we dive deeper in the discussion started in Section \ref{qualitative_insights_main} to provide a more qualitative interpretation of the resulting KBs.

In particular, focusing on babylonGPTKB, we first look at the number of exact-matching triples shared across the 10 runs of the base setting for three entities randomly sampled from those generated by GPT-4.1-mini. The results reported in Tables \ref{tab:shared_triples_1} (for \textit{Hammurabi}), \ref{tab:shared_triples_3} (for \textit{Celestial Bull}), and \ref{tab:shared_triples_4} (for \textit{Ereshkigal}) show that it is rare to find the exact same triples for the same entity across repeated runs. For example, the number of triples for \textit{Hammurabi} ranges from 14 in run 9 to 40 in run 7, in this case with the former 14 all contained in the latter 40. Interestingly, the 7 triples of run 1 for \textit{Celestial Bull} are matched only in run 5, with all remaining runs missing any triple for this entity. Finally, the highest number of triples for \textit{Ereshkigal} is 18 in run 1, but interestingly seven out of the ten runs share the exact same 10 triples. However, it is noteworthy to consider that 1) the reproducibility experiments already show the imperfection of exact-matching metrics in this context, so one can expect a much higher alignment across runs when accounting for semantically similar triples (indeed, among the 1060 entities exclusive to run 1 compared to run 2 based on exact match, 152 do have a semantic match with cosine similarity >=95\% in run 2), and 2) the yield metric across runs is still highly similar across all categories, suggesting that the number of facts for each entity might change, but in general the model's knowledge about that domain is extracted in a pretty stable manner.

Furthermore, we also provide some insights gained from a qualitative comparison of the first GPT run of the base setting against the second run, identifying potential reasons explaining why 20 sample entities are found only in run 1 and not in run 2 based on exact match. The reasons identified can be found in Table \ref{missing_entities}. Overall, the main reason for this mismatch seems to be the lack of the target entity's parent(s) in run 2, representing a missing link in the graph e.g.\ for the entities \textit{Tushratta}, \textit{Yahweh}, or \textit{Shaushtatar}. In other cases, the parent entity is found in both runs, but the target entity is not found in run 2 due to the lack of the predicate which would have lead from the parent to it e.g.\ for the entity \textit{Humbaba}, or either the parent or the target entities are found, but classified differently (e.g.\ the target entity is classified as a literal in run 2) e.g.\ for the entity \textit{god of wisdom}. Finally, we identify 5 of these samples to be cases of simple fuzzy matching i.e.\ the entities are actually found in run 2, just slightly different (e.g.\, different spelling or roman number of a king missing).

Finally, in Table \ref{tbbt_all_entities} we report a sample of entities of runs 1 and 2 in tbbtGPTKB, alphabetically sorted and aligned to showcase overlap and differences.

\begin{table*}[ht]
\small
\resizebox{\textwidth}{!}{%
\begin{tabular}{r|rrrrrrrrrr}
\multicolumn{1}{l}{} & \textbf{Run\_1}            & \textbf{Run\_2}            & \textbf{Run\_3}            & \textbf{Run\_4}            & \textbf{Run\_5}            & \textbf{Run\_6}            & \textbf{Run\_7}            & \textbf{Run\_8}            & \textbf{Run\_9}            & \textbf{Run\_10}           \\
\hline
\textbf{Run\_1}      & \cellcolor[HTML]{C0C0C0}32 & \cellcolor[HTML]{FFFFFF}7  & \cellcolor[HTML]{FFFFFF}14 & \cellcolor[HTML]{FFFFFF}14 & \cellcolor[HTML]{FFFFFF}8  & \cellcolor[HTML]{FFFFFF}11 & \cellcolor[HTML]{FFFFFF}13 & \cellcolor[HTML]{FFFFFF}11 & \cellcolor[HTML]{FFFFFF}10 & \cellcolor[HTML]{FFFFFF}12 \\
\textbf{Run\_2}      & \cellcolor[HTML]{FFFFFF}7  & \cellcolor[HTML]{C0C0C0}33 & \cellcolor[HTML]{FFFFFF}11 & \cellcolor[HTML]{FFFFFF}10 & \cellcolor[HTML]{FFFFFF}3  & \cellcolor[HTML]{FFFFFF}9  & \cellcolor[HTML]{FFFFFF}13 & \cellcolor[HTML]{FFFFFF}9  & \cellcolor[HTML]{FFFFFF}7  & \cellcolor[HTML]{FFFFFF}14 \\
\textbf{Run\_3}      & \cellcolor[HTML]{FFFFFF}14 & \cellcolor[HTML]{FFFFFF}11 & \cellcolor[HTML]{C0C0C0}38 & \cellcolor[HTML]{FFFFFF}13 & \cellcolor[HTML]{FFFFFF}7  & \cellcolor[HTML]{FFFFFF}13 & \cellcolor[HTML]{FFFFFF}18 & \cellcolor[HTML]{FFFFFF}12 & \cellcolor[HTML]{FFFFFF}9  & \cellcolor[HTML]{FFFFFF}14 \\
\textbf{Run\_4}      & \cellcolor[HTML]{FFFFFF}14 & \cellcolor[HTML]{FFFFFF}10 & \cellcolor[HTML]{FFFFFF}13 & \cellcolor[HTML]{C0C0C0}34 & \cellcolor[HTML]{FFFFFF}7  & \cellcolor[HTML]{FFFFFF}14 & \cellcolor[HTML]{FFFFFF}15 & \cellcolor[HTML]{FFFFFF}10 & \cellcolor[HTML]{FFFFFF}10 & \cellcolor[HTML]{FFFFFF}8  \\
\textbf{Run\_5}      & \cellcolor[HTML]{FFFFFF}8  & \cellcolor[HTML]{FFFFFF}3  & \cellcolor[HTML]{FFFFFF}7  & \cellcolor[HTML]{FFFFFF}7  & \cellcolor[HTML]{C0C0C0}17 & \cellcolor[HTML]{FFFFFF}6  & \cellcolor[HTML]{FFFFFF}8  & \cellcolor[HTML]{FFFFFF}7  & \cellcolor[HTML]{FFFFFF}6  & \cellcolor[HTML]{FFFFFF}6  \\
\textbf{Run\_6}      & \cellcolor[HTML]{FFFFFF}11 & \cellcolor[HTML]{FFFFFF}9  & \cellcolor[HTML]{FFFFFF}13 & \cellcolor[HTML]{FFFFFF}14 & \cellcolor[HTML]{FFFFFF}6  & \cellcolor[HTML]{C0C0C0}24 & \cellcolor[HTML]{FFFFFF}17 & \cellcolor[HTML]{FFFFFF}11 & \cellcolor[HTML]{FFFFFF}13 & \cellcolor[HTML]{FFFFFF}10 \\
\textbf{Run\_7}      & \cellcolor[HTML]{FFFFFF}13 & \cellcolor[HTML]{FFFFFF}13 & \cellcolor[HTML]{FFFFFF}18 & \cellcolor[HTML]{FFFFFF}15 & \cellcolor[HTML]{FFFFFF}8  & \cellcolor[HTML]{FFFFFF}17 & \cellcolor[HTML]{C0C0C0}40 & \cellcolor[HTML]{FFFFFF}11 & \cellcolor[HTML]{FFFFFF}14 & \cellcolor[HTML]{FFFFFF}14 \\
\textbf{Run\_8}      & \cellcolor[HTML]{FFFFFF}11 & \cellcolor[HTML]{FFFFFF}9  & \cellcolor[HTML]{FFFFFF}12 & \cellcolor[HTML]{FFFFFF}10 & \cellcolor[HTML]{FFFFFF}7  & \cellcolor[HTML]{FFFFFF}11 & \cellcolor[HTML]{FFFFFF}11 & \cellcolor[HTML]{C0C0C0}20 & \cellcolor[HTML]{FFFFFF}8  & \cellcolor[HTML]{FFFFFF}8  \\
\textbf{Run\_9}      & \cellcolor[HTML]{FFFFFF}10 & \cellcolor[HTML]{FFFFFF}7  & \cellcolor[HTML]{FFFFFF}9  & \cellcolor[HTML]{FFFFFF}10 & \cellcolor[HTML]{FFFFFF}6  & \cellcolor[HTML]{FFFFFF}13 & \cellcolor[HTML]{FFFFFF}14 & \cellcolor[HTML]{FFFFFF}8  & \cellcolor[HTML]{C0C0C0}14 & \cellcolor[HTML]{FFFFFF}7  \\
\textbf{Run\_10}     & \cellcolor[HTML]{FFFFFF}12 & \cellcolor[HTML]{FFFFFF}14 & \cellcolor[HTML]{FFFFFF}14 & \cellcolor[HTML]{FFFFFF}8  & \cellcolor[HTML]{FFFFFF}6  & \cellcolor[HTML]{FFFFFF}10 & \cellcolor[HTML]{FFFFFF}14 & \cellcolor[HTML]{FFFFFF}8  & \cellcolor[HTML]{FFFFFF}7  & \cellcolor[HTML]{C0C0C0}31
\end{tabular}}
\caption{Triples shared across the 10 runs of the base setting in babylonGPTKB for the seed entity \textit{Hammurabi}.}
\label{tab:shared_triples_1}
\end{table*}

\begin{table*}[ht]
\small
\resizebox{\textwidth}{!}{%
\begin{tabular}{r|rrrrrrrrrr}
\multicolumn{1}{l}{} & \textbf{Run\_1}           & \textbf{Run\_2}           & \textbf{Run\_3}           & \textbf{Run\_4}           & \textbf{Run\_5}           & \textbf{Run\_6}           & \textbf{Run\_7}           & \textbf{Run\_8}           & \textbf{Run\_9}           & \textbf{Run\_10}          \\
\hline
\textbf{Run\_1}      & \cellcolor[HTML]{C0C0C0}7 & \cellcolor[HTML]{FFFFFF}0 & \cellcolor[HTML]{FFFFFF}0 & \cellcolor[HTML]{FFFFFF}0 & \cellcolor[HTML]{FFFFFF}7 & \cellcolor[HTML]{FFFFFF}0 & \cellcolor[HTML]{FFFFFF}0 & \cellcolor[HTML]{FFFFFF}0 & \cellcolor[HTML]{FFFFFF}0 & \cellcolor[HTML]{FFFFFF}0 \\
\textbf{Run\_2}      & \cellcolor[HTML]{FFFFFF}0 & \cellcolor[HTML]{C0C0C0}0 & \cellcolor[HTML]{FFFFFF}0 & \cellcolor[HTML]{FFFFFF}0 & \cellcolor[HTML]{FFFFFF}0 & \cellcolor[HTML]{FFFFFF}0 & \cellcolor[HTML]{FFFFFF}0 & \cellcolor[HTML]{FFFFFF}0 & \cellcolor[HTML]{FFFFFF}0 & \cellcolor[HTML]{FFFFFF}0 \\
\textbf{Run\_3}      & \cellcolor[HTML]{FFFFFF}0 & \cellcolor[HTML]{FFFFFF}0 & \cellcolor[HTML]{C0C0C0}0 & \cellcolor[HTML]{FFFFFF}0 & \cellcolor[HTML]{FFFFFF}0 & \cellcolor[HTML]{FFFFFF}0 & \cellcolor[HTML]{FFFFFF}0 & \cellcolor[HTML]{FFFFFF}0 & \cellcolor[HTML]{FFFFFF}0 & \cellcolor[HTML]{FFFFFF}0 \\
\textbf{Run\_4}      & \cellcolor[HTML]{FFFFFF}0 & \cellcolor[HTML]{FFFFFF}0 & \cellcolor[HTML]{FFFFFF}0 & \cellcolor[HTML]{C0C0C0}0 & \cellcolor[HTML]{FFFFFF}0 & \cellcolor[HTML]{FFFFFF}0 & \cellcolor[HTML]{FFFFFF}0 & \cellcolor[HTML]{FFFFFF}0 & \cellcolor[HTML]{FFFFFF}0 & \cellcolor[HTML]{FFFFFF}0 \\
\textbf{Run\_5}      & \cellcolor[HTML]{FFFFFF}7 & \cellcolor[HTML]{FFFFFF}0 & \cellcolor[HTML]{FFFFFF}0 & \cellcolor[HTML]{FFFFFF}0 & \cellcolor[HTML]{C0C0C0}7 & \cellcolor[HTML]{FFFFFF}0 & \cellcolor[HTML]{FFFFFF}0 & \cellcolor[HTML]{FFFFFF}0 & \cellcolor[HTML]{FFFFFF}0 & \cellcolor[HTML]{FFFFFF}0 \\
\textbf{Run\_6}      & \cellcolor[HTML]{FFFFFF}0 & \cellcolor[HTML]{FFFFFF}0 & \cellcolor[HTML]{FFFFFF}0 & \cellcolor[HTML]{FFFFFF}0 & \cellcolor[HTML]{FFFFFF}0 & \cellcolor[HTML]{C0C0C0}0 & \cellcolor[HTML]{FFFFFF}0 & \cellcolor[HTML]{FFFFFF}0 & \cellcolor[HTML]{FFFFFF}0 & \cellcolor[HTML]{FFFFFF}0 \\
\textbf{Run\_7}      & \cellcolor[HTML]{FFFFFF}0 & \cellcolor[HTML]{FFFFFF}0 & \cellcolor[HTML]{FFFFFF}0 & \cellcolor[HTML]{FFFFFF}0 & \cellcolor[HTML]{FFFFFF}0 & \cellcolor[HTML]{FFFFFF}0 & \cellcolor[HTML]{C0C0C0}0 & \cellcolor[HTML]{FFFFFF}0 & \cellcolor[HTML]{FFFFFF}0 & \cellcolor[HTML]{FFFFFF}0 \\
\textbf{Run\_8}      & \cellcolor[HTML]{FFFFFF}0 & \cellcolor[HTML]{FFFFFF}0 & \cellcolor[HTML]{FFFFFF}0 & \cellcolor[HTML]{FFFFFF}0 & \cellcolor[HTML]{FFFFFF}0 & \cellcolor[HTML]{FFFFFF}0 & \cellcolor[HTML]{FFFFFF}0 & \cellcolor[HTML]{C0C0C0}0 & \cellcolor[HTML]{FFFFFF}0 & \cellcolor[HTML]{FFFFFF}0 \\
\textbf{Run\_9}      & \cellcolor[HTML]{FFFFFF}0 & \cellcolor[HTML]{FFFFFF}0 & \cellcolor[HTML]{FFFFFF}0 & \cellcolor[HTML]{FFFFFF}0 & \cellcolor[HTML]{FFFFFF}0 & \cellcolor[HTML]{FFFFFF}0 & \cellcolor[HTML]{FFFFFF}0 & \cellcolor[HTML]{FFFFFF}0 & \cellcolor[HTML]{C0C0C0}0 & \cellcolor[HTML]{FFFFFF}0 \\
\textbf{Run\_10}     & \cellcolor[HTML]{FFFFFF}0 & \cellcolor[HTML]{FFFFFF}0 & \cellcolor[HTML]{FFFFFF}0 & \cellcolor[HTML]{FFFFFF}0 & \cellcolor[HTML]{FFFFFF}0 & \cellcolor[HTML]{FFFFFF}0 & \cellcolor[HTML]{FFFFFF}0 & \cellcolor[HTML]{FFFFFF}0 & \cellcolor[HTML]{FFFFFF}0 & \cellcolor[HTML]{C0C0C0}0
\end{tabular}}
\caption{Triples shared across the 10 runs of the base setting in babylonGPTKB for the seed entity \textit{Celestial Bull}.}
\label{tab:shared_triples_3}
\end{table*}

\begin{table*}[ht]
\small
\resizebox{\textwidth}{!}{%
\begin{tabular}{r|rrrrrrrrrr}
\multicolumn{1}{l}{} & \textbf{Run\_1}            & \textbf{Run\_2}            & \textbf{Run\_3}            & \textbf{Run\_4}            & \textbf{Run\_5}            & \textbf{Run\_6}            & \textbf{Run\_7}            & \textbf{Run\_8}            & \textbf{Run\_9}            & \textbf{Run\_10}           \\
\hline
\textbf{Run\_1}      & \cellcolor[HTML]{C0C0C0}18 & \cellcolor[HTML]{FFFFFF}4  & \cellcolor[HTML]{FFFFFF}8  & \cellcolor[HTML]{FFFFFF}10 & \cellcolor[HTML]{FFFFFF}10 & \cellcolor[HTML]{FFFFFF}10 & \cellcolor[HTML]{FFFFFF}10 & \cellcolor[HTML]{FFFFFF}10 & \cellcolor[HTML]{FFFFFF}10 & \cellcolor[HTML]{FFFFFF}11 \\
\textbf{Run\_2}      & \cellcolor[HTML]{FFFFFF}4  & \cellcolor[HTML]{C0C0C0}10 & \cellcolor[HTML]{FFFFFF}6  & \cellcolor[HTML]{FFFFFF}4  & \cellcolor[HTML]{FFFFFF}4  & \cellcolor[HTML]{FFFFFF}4  & \cellcolor[HTML]{FFFFFF}4  & \cellcolor[HTML]{FFFFFF}4  & \cellcolor[HTML]{FFFFFF}4  & \cellcolor[HTML]{FFFFFF}4  \\
\textbf{Run\_3}      & \cellcolor[HTML]{FFFFFF}8  & \cellcolor[HTML]{FFFFFF}6  & \cellcolor[HTML]{C0C0C0}10 & \cellcolor[HTML]{FFFFFF}8  & \cellcolor[HTML]{FFFFFF}8  & \cellcolor[HTML]{FFFFFF}8  & \cellcolor[HTML]{FFFFFF}8  & \cellcolor[HTML]{FFFFFF}8  & \cellcolor[HTML]{FFFFFF}8  & \cellcolor[HTML]{FFFFFF}8  \\
\textbf{Run\_4}      & \cellcolor[HTML]{FFFFFF}10 & \cellcolor[HTML]{FFFFFF}4  & \cellcolor[HTML]{FFFFFF}8  & \cellcolor[HTML]{C0C0C0}10 & \cellcolor[HTML]{FFFFFF}10 & \cellcolor[HTML]{FFFFFF}10 & \cellcolor[HTML]{FFFFFF}10 & \cellcolor[HTML]{FFFFFF}10 & \cellcolor[HTML]{FFFFFF}10 & \cellcolor[HTML]{FFFFFF}10 \\
\textbf{Run\_5}      & \cellcolor[HTML]{FFFFFF}10 & \cellcolor[HTML]{FFFFFF}4  & \cellcolor[HTML]{FFFFFF}8  & \cellcolor[HTML]{FFFFFF}10 & \cellcolor[HTML]{C0C0C0}10 & \cellcolor[HTML]{FFFFFF}10 & \cellcolor[HTML]{FFFFFF}10 & \cellcolor[HTML]{FFFFFF}10 & \cellcolor[HTML]{FFFFFF}10 & \cellcolor[HTML]{FFFFFF}10 \\
\textbf{Run\_6}      & \cellcolor[HTML]{FFFFFF}10 & \cellcolor[HTML]{FFFFFF}4  & \cellcolor[HTML]{FFFFFF}8  & \cellcolor[HTML]{FFFFFF}10 & \cellcolor[HTML]{FFFFFF}10 & \cellcolor[HTML]{C0C0C0}12 & \cellcolor[HTML]{FFFFFF}10 & \cellcolor[HTML]{FFFFFF}10 & \cellcolor[HTML]{FFFFFF}10 & \cellcolor[HTML]{FFFFFF}11 \\
\textbf{Run\_7}      & \cellcolor[HTML]{FFFFFF}10 & \cellcolor[HTML]{FFFFFF}4  & \cellcolor[HTML]{FFFFFF}8  & \cellcolor[HTML]{FFFFFF}10 & \cellcolor[HTML]{FFFFFF}10 & \cellcolor[HTML]{FFFFFF}10 & \cellcolor[HTML]{C0C0C0}10 & \cellcolor[HTML]{FFFFFF}10 & \cellcolor[HTML]{FFFFFF}10 & \cellcolor[HTML]{FFFFFF}10 \\
\textbf{Run\_8}      & \cellcolor[HTML]{FFFFFF}10 & \cellcolor[HTML]{FFFFFF}4  & \cellcolor[HTML]{FFFFFF}8  & \cellcolor[HTML]{FFFFFF}10 & \cellcolor[HTML]{FFFFFF}10 & \cellcolor[HTML]{FFFFFF}10 & \cellcolor[HTML]{FFFFFF}10 & \cellcolor[HTML]{C0C0C0}10 & \cellcolor[HTML]{FFFFFF}10 & \cellcolor[HTML]{FFFFFF}10 \\
\textbf{Run\_9}      & \cellcolor[HTML]{FFFFFF}10 & \cellcolor[HTML]{FFFFFF}4  & \cellcolor[HTML]{FFFFFF}8  & \cellcolor[HTML]{FFFFFF}10 & \cellcolor[HTML]{FFFFFF}10 & \cellcolor[HTML]{FFFFFF}10 & \cellcolor[HTML]{FFFFFF}10 & \cellcolor[HTML]{FFFFFF}10 & \cellcolor[HTML]{C0C0C0}10 & \cellcolor[HTML]{FFFFFF}10 \\
\textbf{Run\_10}     & \cellcolor[HTML]{FFFFFF}11 & \cellcolor[HTML]{FFFFFF}4  & \cellcolor[HTML]{FFFFFF}8  & \cellcolor[HTML]{FFFFFF}10 & \cellcolor[HTML]{FFFFFF}10 & \cellcolor[HTML]{FFFFFF}11 & \cellcolor[HTML]{FFFFFF}10 & \cellcolor[HTML]{FFFFFF}10 & \cellcolor[HTML]{FFFFFF}10 & \cellcolor[HTML]{C0C0C0}16
\end{tabular}}
\caption{Triples shared across the 10 runs of the base setting in babylonGPTKB for the seed entity \textit{Ereshkigal}.}
\label{tab:shared_triples_4}
\end{table*}

\clearpage

\begin{table*}[h]
\begin{tabular}{p{5.7cm}|p{3.5cm}|p{5cm}}
\hline
\textbf{Entity} & \textbf{Reason} & \textbf{Notes} \\ 
\hline
Tushratta & missing parent & all three parents missing \\
Samos & missing parent &  \\
Mordecai & missing parent &  \\
Shaushtatar & missing parent &  \\
Yahweh & missing parent &  \\
Huzziya I & missing parent & missing parent, but shared grandparent, for which there are different predicates \\
Rav Papa & missing parent & missing also grandparent \\
Babylonian jewish academies & missing parent & both parents missing \\
Ur-Lumma & missing parent & both parents missing \\
Valley of the Kings & missing parent & both parents missing \\
\hline
Shutruk-Nahhunte & fuzzy matching & different spelling "King Shutruk-Nakhkhunte" \\
Shu-Ilishu I & fuzzy matching & roman number missing \\
City of Hillah & fuzzy matching & mentioned as "modern city of Hillah" \\
God Marduk & fuzzy matching & mentioned as "Marduk" \\
exile of Israelites & fuzzy matching & mentioned as "Exile of Jewish population" \\
\hline
Humbaba & missing predicate & one out of two parents found \\
shrine to marduk & missing predicate &  \\
Tishpak & missing predicate & parent "City of Eshnunna" not found exactly like that, but found as "Eshnunna" \\
\hline
god of wisdom & classified differently & in run 1 NE, in run 2 literal \\
Leader of the ancient city of Babylon & classified differently & in run 1 NE, in run 2 literal \\
\hline
\end{tabular}
\caption{20 entities randomly sampled from run 1 of the base setting, not found in run 2, with respective reasons explaining the mismatch.}
\label{missing_entities}
\end{table*}

\begin{table*}
\begin{tabular}{c|c}
\hline
\textbf{Run 1} & \textbf{Run 2} \\
\hline
-                                                                  & A turtle named Sheldon Jr.                                                             \\
Aarti Mann                                                         & -                                                                                      \\
-                                                                  & Academy of Television Arts \& Sciences                                                 \\
Alfred Hofstadter                                                  & Alfred Hofstadter                                                                      \\
-                                                                  & Alfred Hofstadter Jr.                                                                  \\
Allegra Hofstadter                                                 & -                                                                                      \\
Amy Farrah Fowler                                                  & Amy Farrah Fowler                                                                      \\
-                                                                  & Amy and Sheldon start dating                                                           \\
Anthony Rich                                                       & Anthony Rich                                                                           \\
Anu                                                                & Anu                                                                                    \\
-                                                                  & Apartment 4A                                                                           \\
-                                                                  & Apartment with Rajesh Koothrappali                                                     \\
-                                                                  & Arthur Jeffries                                                                        \\
Barenaked Ladies                                                   & Barenaked Ladies                                                                       \\
-                                                                  & Barry Kripke                                                                           \\
-                                                                  & Bazinga                                                                                \\
Bazinga!                                                           & Bazinga!                                                                               \\
Bernadette Rostenkowski                                            & Bernadette Rostenkowski                                                                \\
Bernadette Rostenkowski-Wolowitz                                   & Bernadette Rostenkowski-Wolowitz                                                       \\
-                                                                  & Bernadette Rostenkowski-Wolowitz (stepmother)                                          \\
-                                                                  & Bernadette Rostenkowski-Wolowitz (wife)                                                \\
-                                                                  & Bernadette Wolowitz                                                                    \\
-                                                                  & Bert Kibbler                                                                           \\
Beverly Hofstadter                                                 & Beverly Hofstadter                                                                     \\
-                                                                  & Beverly Hofstadter (ex-wife)                                                           \\
Bill Prady                                                         & Bill Prady                                                                             \\
-                                                                  & Billy Gardell                                                                          \\
-                                                                  & Bob Newhart                                                                            \\
-                                                                  & Brian Posehn                                                                           \\
Broadcast Television Journalists Association                       & -                                                                                      \\
CBS                                                                & CBS                                                                                    \\
-                                                                  & California                                                                             \\
Caltech                                                            & Caltech                                                                                \\
Caltech Researcher                                                 & -                                                                                      \\
-                                                                  & Caltech Theoretical Physicist                                                          \\
-                                                                  & Cambridge, Massachusetts                                                               \\
-                                                                  & Carl Reiner                                                                            \\
Carol Ann Susi                                                     & Carol Ann Susi                                                                         \\
-                                                                  & Casey Sander                                                                           \\
-                                                                  & Christianity                                                                           \\
Christine Baranski                                                 & Christine Baranski                                                                     \\
Chuck Lorre                                                        & Chuck Lorre                                                                            \\
Chuck Lorre Productions                                            & Chuck Lorre Productions                                                                \\
Cinnamon (dog)                                                     & -                                                                                      \\
-                                                                  & Cinnamon the dog       \\         
\hline
\end{tabular}
\caption{Sampled entities of runs 1 and 2 in tbbtGPTKB, alphabetically sorted and aligned to showcase overlap and differences.}
\label{tbbt_all_entities}
\end{table*}

\clearpage

\section{Hardware specifications for running local models}
\label{app:hardware_specs}
The experiments with local models were conducted using a local HPC system, serving the LLMs with the following GPU hardware specifications:

\begin{itemize}
    \item \textbf{gpt-oss-120b}: 1x H100
    \item \textbf{Llama 3.3 70b}: 4x A100
    \item \textbf{Llama 4 Scout}: 4x H100
    \item \textbf{DeepSeek-R1}: 8x H100
    \item \textbf{Teuken 7b Instruct}: 1x A100
\end{itemize}

\section{Plots and results of the ensembling for final KBs}
\label{app:ensembling}

\begin{figure}[H]
    \centering
    \includegraphics[width=1\linewidth]{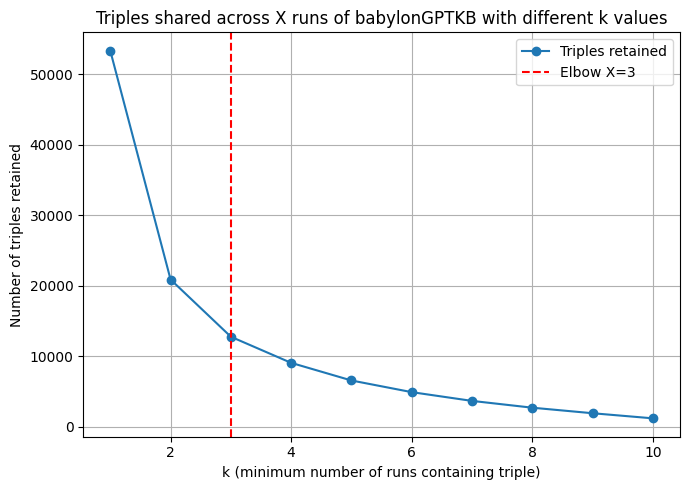}
    \caption{Triples shared across X runs of babylonGPTKB with different k values.}
    \label{fig:elbow_babylon}
\end{figure}

\begin{figure}[H]
    \centering
    \includegraphics[width=1\linewidth]{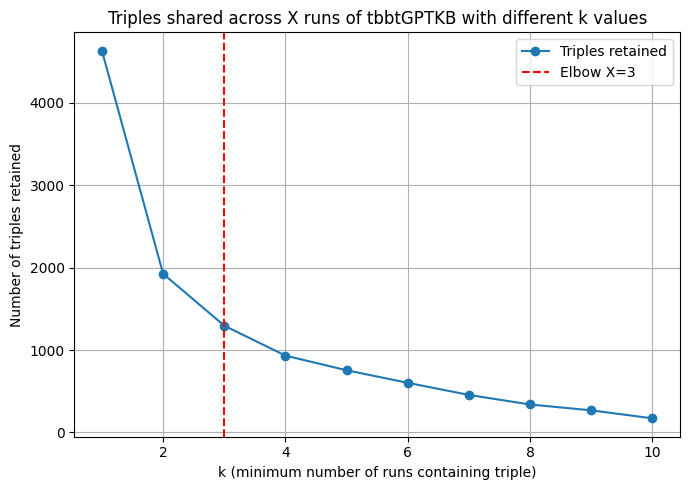}
    \caption{Triples shared across X runs of tbbtGPTKB with different k values.}
    \label{fig:elbow_tbbt}
\end{figure}

\begin{figure}[H]
    \centering
    \includegraphics[width=1\linewidth]{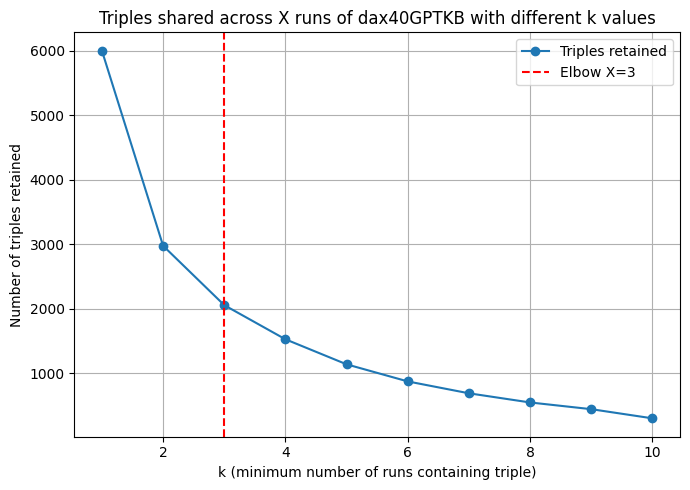}
    \caption{Triples shared across X runs of dax40GPTKB with different k values.}
    \label{fig:elbow_dax40}
\end{figure}

In Figures \ref{fig:elbow_babylon}, \ref{fig:elbow_tbbt}, and \ref{fig:elbow_dax40} we report the plots of triples shared across X of each topic's base setting runs against different numbers of \textit{k}. Each plot shows the ``elbow point'' equals 3. The elbow method is a heuristic which involves plotting a variable (the number of shared triples) against \textit{k} and selecting the point farthest from a straight line drawn between the two extreme points in the plot (i.e.\ where the curve drops sharply, forming an “elbow” in the curve). Furthermore, in Table \ref{tab:elbow}, we report the actual number of triples at different \textit{k} values. In bold the elbow point.

\begin{table}[H]
    \centering
    \begin{tabular}{c|c|c|c}
    \hline
        \textbf{\textit{k} runs} & \textbf{Babylon} & \textbf{TBBT} & \textbf{DAX 40} \\
        \hline
        1 & 53,320 & 4631 & 5999\\
        2 & 20,859 & 1925 & 2976\\
        \textbf{3} & \textbf{12,738} & \textbf{1295} & \textbf{2056}\\
        4 & 9033 & 932 & 1527\\
        5 & 6532 & 754 & 1139\\
        6 & 4877 & 602 & 874\\
        7 & 3635 & 455 & 690\\
        8 & 2670 & 339 & 548\\
        9 & 1884 & 268 & 444\\
        10 & 1160 & 171 & 302\\
        \hline
    \end{tabular}
    \caption{Number of triples retained per topic found in at least \textit{k} runs of each topic's base setting.}
    \label{tab:elbow}
\end{table}

\clearpage

\section{Manual accuracy evaluation}
\label{acc_eval}
We conduct a small-scale accuracy evaluation using human judges by sampling 200 random triples from babylonGPTKB (Table \ref{tab:acc_eval}). The first half of the sample is extracted from the first reproducibility run, while the second half is drawn from the ensembled KB released publicly. Our evaluation shows an accuracy of 93\% for the first group, with 4\% of triples judged as false and 3\% as unverifiable, and 95\% for the second group with 3\% false and 2\% unverifiable triples.

While these numbers suggest that the extracted triples are generally reliable, the limited sample size and the presence of unverifiable cases mean that these results should be interpreted as indicative rather than conclusive. As stated in the Limitations of this work, the evaluation of factual correctness was not the primary objective of our experiments, therefore broader evaluations would be needed to fully characterize accuracy.

\begin{table}[h]
    \centering
    \begin{tabular}{c|c|c|c}
    \hline
         & \textbf{True} & \textbf{False} & \textbf{Unverifiable} \\
         \hline
        \textbf{1st repr. run} & 93 & 4 & 3\\
        \textbf{Released KB} & 95 & 3 & 2\\
        \textbf{TOTAL} & 188 & 7 & 5\\
        \hline
    \end{tabular}
    \caption{Results of the small-scale accuracy evaluation of babylonGPTKB.}
    \label{tab:acc_eval}
\end{table}

\section{Additional Visualizations}
\label{app:additional_visualizations}

\begin{figure*}[] 
    \centering
    \includegraphics[width=\textwidth]{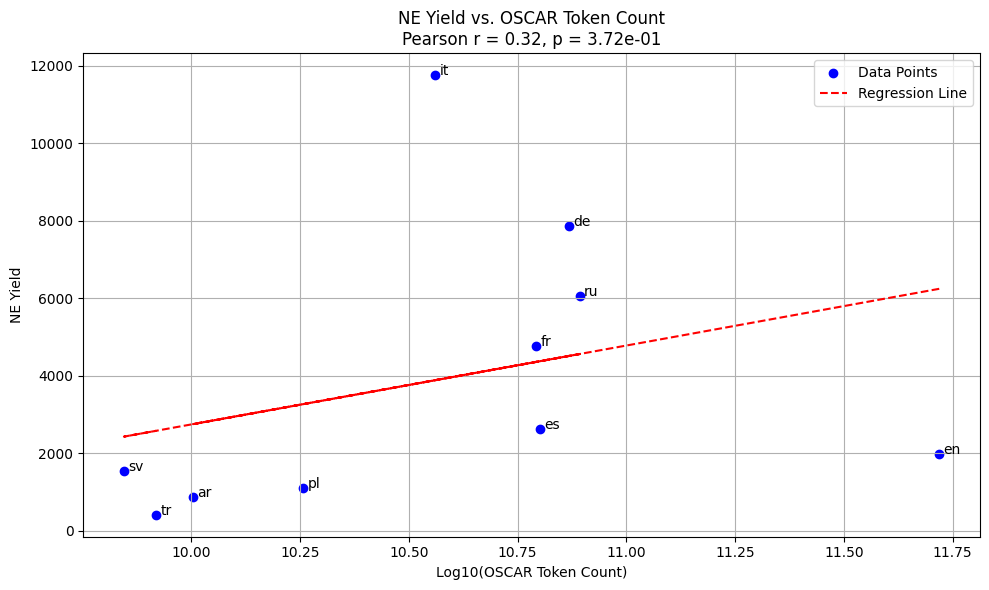}
    \caption{Named entities yielded per language in our experiments compared to language distribution over the web (quantified in the token count statistics of the OSCAR web crawl corpus \cite{OrtizSuarezSagotRomary2019}). The correlation is weak and not statistically significant.}
    \label{fig:oscar}
\end{figure*}

\begin{figure*}[]
    \centering
    \includegraphics[width=\textwidth]{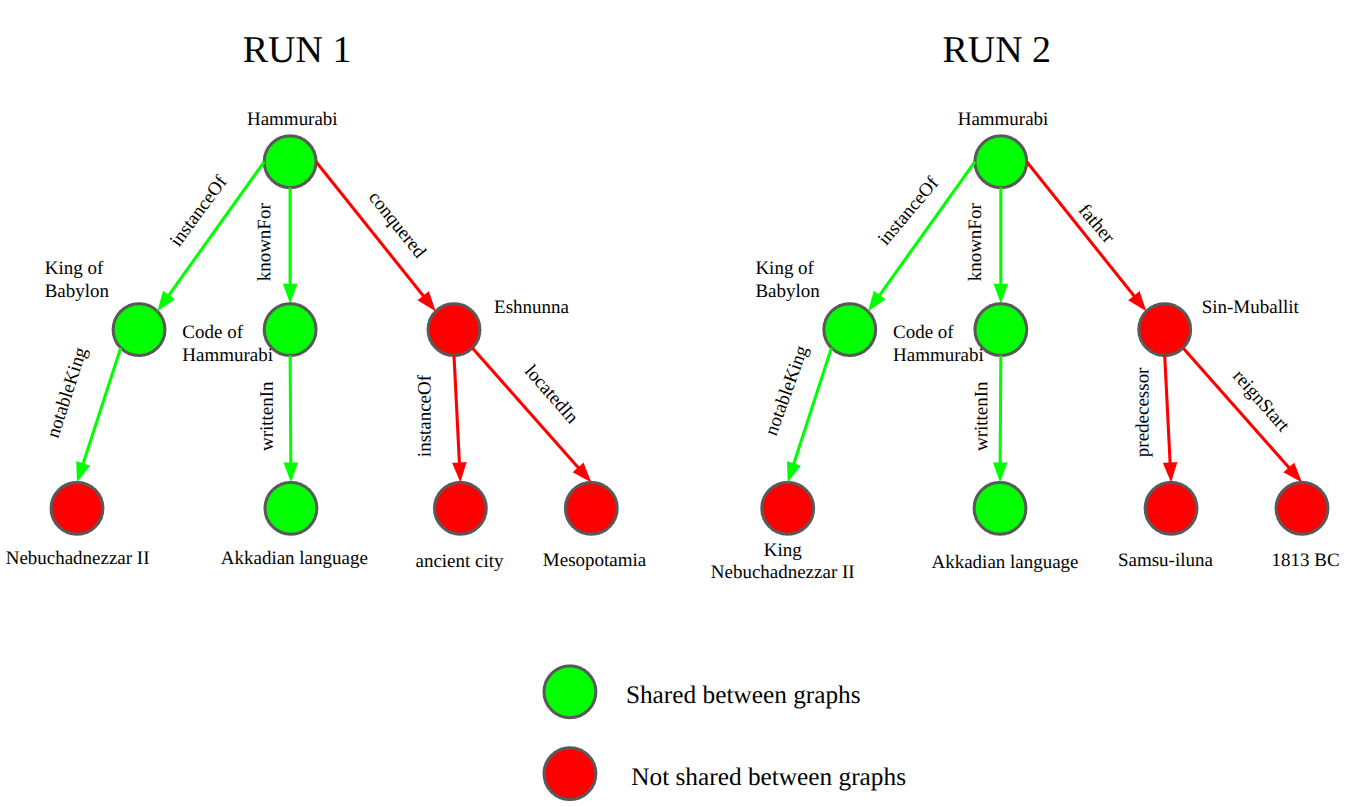}
    \caption{Illustrative visualization of the first two layers of babylonGPTKB in the 1st and 2nd run. Notice overlap and differences.}
    \label{fig:kg_viz_layer1-2}
\end{figure*}

\end{document}